\documentclass[journal,transmag]{IEEEtran}
\ifCLASSINFOpdf
   \usepackage[pdftex]{graphicx}
\else
\fi

\begin{document}
%
\title{Classifying Object Manipulation Actions based on Grasp-types and Motion-Constraints}


\author{\IEEEauthorblockN{Kartik Gupta\IEEEauthorrefmark{1},
Darius Burschka\IEEEauthorrefmark{2}, and
Arnav Bhavsar\IEEEauthorrefmark{3}}\\
\IEEEauthorblockA{\IEEEauthorrefmark{1}Australian National University, Australia}
\IEEEauthorblockA{\IEEEauthorrefmark{2}Technical University Munich, Germany}
\IEEEauthorblockA{\IEEEauthorrefmark{3}Indian Institute of Technology Mandi, India}\\
}

%



\maketitle
\begin{abstract}
In this work, we address a challenging problem of fine-grained and coarse-grained recognition of object manipulation actions. Due to the variations in geometrical and motion constraints, there are different manipulations actions possible to perform different sets of actions with an object. Also, there are subtle movements involved to complete most of object manipulation actions. This makes the task of object manipulation action recognition difficult with only just the motion information. We propose to use grasp and motion-constraints information to recognise and understand action intention with different objects. We also provide an extensive experimental evaluation on the recent Yale Human Grasping dataset consisting of large set of 455 manipulation actions. The evaluation involves a) Different contemporary multi-class classifiers, and binary classifiers with one-vs-one multi-class voting scheme, b) Differential comparisons results based on subsets of attributes involving information of grasp and motion-constraints, c) Fine-grained and Coarse-grained object manipulation action recognition based on fine-grained as well as coarse-grained grasp type information, and d) Comparison between Instance level and Sequence level modeling of object manipulation actions. Our results justifies the efficacy of grasp attributes for the task of fine-grained and coarse-grained object manipulation action recognition.
\end{abstract}




%
\IEEEpeerreviewmaketitle

\section{Introduction}
Human action recognition for full body parts has been studied in \cite{zhu2016}, \cite{poppe} for various applications such as video surveillance, content-based video search, and human-robot interactions. Most studies consider the aspect of short lived actions, where the beginning and end of the actions is explicitly specified. Later efforts have been made to the recognition of movements along with the associated objects, both problems of great interest to the study of action analysis. Still, these methods are less reliable for the case of manipulation actions which are performed at finer level. In the household and work environment tasks, considering actions involving local body parts, is important. The reason for this is the slight movement of hands required to accomplish most of these tasks, and these hand movements are not clearly perceivable through motion information from sensors. This is crucially important towards modeling and monitoring the behavior of the individuals, and also in transferring the object manipulation capabilities to the robots for performing both the household and workforce tasks. Such action recognition based technologies can also benefit various domains such as entertainment, smart homes, elderly care, health rehabilitation, analyzing productivity of human work-tasks etc. 

Human object interactions are largely co-related to the actions performed using a particular object. Although action recognition for the actions specific to the objects is a problem which has been studied in some works \cite{filip}, \cite{jell}, the recognition and understanding of the varied object manipulation actions is still largely unresolved. Everyday manipulation tasks include considerable amount of variations in a particular task being performed. Same action can be performed in varied ways according to the habit and styles of the subject. Most of the manipulation actions contain very subtle variations in the observed whole body motion trajectories for the action being performed. These factors make the problem of recognizing a large set  of manipulation tasks, a challenging job.

\begin{figure}[!t]
  \centering
  \includegraphics[width=9cm]{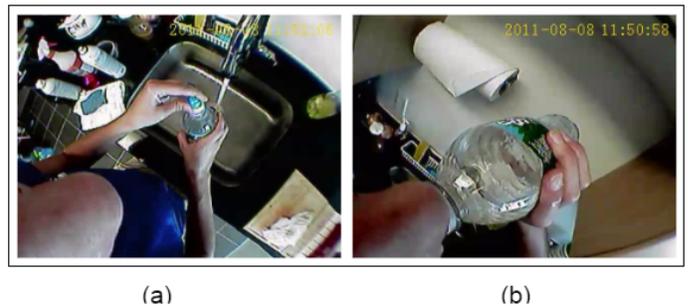}
  \caption{Instances from Yale human grasping dataset depicting (a) Precision grasp for opening the bottle and (b) Power grasp for drinking.}
  \label{bottle}
  \vspace{-0.5cm}
\end{figure}

Most of the action recognition frameworks are specifically designed for a smaller and specific set of actions as the motion dynamics based modeling are trained specifically to distinguish them from the actions falling in that specific action set. While motion dynamics is important, 
it may not uniquely represent manipulation actions. For example, brushing the teeth and drinking water both have similar sort of movements at a coarse level if take the whole human skeleton instead of hand pose specifically. Thus, it is clear that all actions consisting of picking an object and then interaction of object with mouth and finally releasing an object, will not be different from one another based on motion dynamics. So, there is any inherent requirement other aspects that augment the motion dynamics to have better inter-class differentiation for actions. 

The manipulation actions performed by humans can also be co-related to the hand grasps used to perform the specific actions with an object. This hypothesis is based on the fact that the object manipulation actions are initiated from the point when the first object is grasped at first. Thus, hand grasps types also aids in the segmentation of video sequences temporally, for object manipulation sequences. Human actions can be described at different levels of abstraction and the actions at lower level consists of multiple sub-actions where an object is first grasped, then manipulated and finally released. In most of the actions, the point of initiation is same as the action manipulation and thus a single grasp type is uniquely co-related to the action being performed. However, in some actions there may be multiple grasps being involved for specific type of action, which requires sequential modeling of the grasp types. Another interesting point to note here is that these grasp types are much easier to capture through the normal RGB images in comparison to the motion dynamics.    
   
We propose and evaluate different approaches to utilize the grasp and object motion-constraints based information for fine-grained and coarse-grained recognition of everyday manipulation actions. These actions are performed in workforce environment and household environment by workers trained over years of experience performing these tasks (which allows to better evaluate the generalisations). 
We believe that the ability of classification of object manipulation actions using local body configurations (aspects of grasps) and motion information can allow a good-quality automated recognition of larger set of everyday actions because in general it allows to define the properties specifically unique to these actions. This  grasp based action recognition is essentially more appropriate for the objects which can be manipulated in different ways for different actions. For instance, a particular object (e.g. bottle) can be opened using a precision grasp and can also be used for drinking with power grasp as illustrated in Figure \ref{bottle}. This type of classification allows to identify very useful information about the task intended by the user based on the grasp information, thus also facilitates to the prediction of actions in the scenarios where interactions between the humans and robots is required. As we focus on the task of action recognition, we assume that the information about grasp attributes and motion-constraints is available to us, as in the case of Yale human grasping dataset \cite{yale}, which we have used in this paper.


\section{Related work}
Most of the action recognition methodologies models the action using full-body motion based features, which only works well for the specific class of action recognition problem where action set is relatively small such as in \cite{gupta2016scale}, \cite{vemulapalli}, \cite{sharaf}. These approach do not look useful when it comes to their application on real everyday actions. Research in the area of human action recognition has been mainly focused on full-body motions that can be characterized by movement and change of posture like walking, waving, etc. 

In many action recognition approaches \cite{bobick}, \cite{rao}, \cite{sullivan}, human motion information have been used. The problem of action recognition has been dealt using motion trajectories with the use of depth cameras like Kinect. These approaches (e.g. see \cite{chen}) are typically considered to be more robust to generate human pose information which can be used for the purpose of action recognition. However, Kinect body pose recognition is not accurate when there are human-object interactions due to occlusions. Motion dynamics based action recognition still cannot capture the representation for the subtle object manipulations. Another interesting aspect is the variations in goal of the task with similar motion dynamics. 

Hand gesture recognition is more closer to the problem of object manipulation action recognition. Hand gesture recognition has also been addressed using depth data generated from Kinect in Kurakin et al. \cite{kurakin} and Wang et al. \cite{wang}. But these techniques mainly target sign language gestures and not the human hand-object interactions. Wang et al. \cite{wang} treats an action sequence as a 4D shape and propose random occupancy pattern (ROP) features, extracted from random sampling of 4D subvolumes with different sizes and at different locations. In gesture depth sequences, the semantics of the gestures are mainly understood by the large movement of the hand. These approaches use cropped portion of hand using some hand detection approach, to determine these large hand movements to model different gestures. But, these clear motion information are not easily perceivable in the case object manipulation tasks. 

At this point, we note here that the above mentioned works involve processing low-level information (e.g. feature extraction from videos/images), whereas our goal in this work is to convey the importance of grasp and motion-constraints information at the higher semantic level (e.g. types of grasps and motion-constraints). Such high level attributes for manipulating actions, are indeed available \cite{feixobject}, \cite{feixtask}, \cite{feix} as a part of the Yale human grasping dataset that we are considering in this work. 

The problem of understanding manipulation actions is of great interest in robotics as well, where the focus is on simplifying methods to implement action execution on robots. There has been considerable amount of work in robot task planning based on imitation learning \cite{argall}, which is essentially the problem of object manipulation through robots by imitating the real world trajectory observed on people performing the action. Understanding the specific types of grasp required in the action sequence aids to the purpose of imitation learning as well. The knowledge about how to grasp the object is significant, so the robot can accordingly position its effectors. For example, humanoid robot with one parallel gripper and one vacuum gripper, should select the vacuum gripper for power grasp, but when a precision grasp is needed, the parallel gripper is a better choice. Yang et al. \cite{yang2015robot} presents a system that learns manipulation action plans by processing unconstrained videos from the World Wide Web. It understands the objects and hand grasp types using CNNs (convolutional neural networks) and later finds the candidate actions that can be performed using the recognized objects from trained language model. Finally, they provide an action tree which can be reversely parsed for action execution by robot. 

To the best of our knowledge, apart from \cite{yezhou}, \cite{yang2015robot} and \cite{yang2014cognitive}, there has been no work using grasp information for action recognition. Yang et al. \cite{yezhou} semantically group action intentions using grasp based information into three coarse and somewhat abstract classes: Force-oriented, Skill-oriented, and Casual actions. They use hand grasps recognized through convolutional neural network to understand the class of action, each image belong to. Yang et al. \cite{yang2014cognitive} develop a grammatical formalism for parsing and interpreting action sequences. Their basic idea is to divide actions into sub-actions of when the object is grasped and released, or if there is change in the grasp type during the course of an action motion. This grammatical formalism provides a syntax and semantics of action, over which basic tools for understanding of actions can be developed. Feix et al. \cite{haptics} considers the problem of grasp classification on Yale human grasping dataset, again based on the coarsely defined task attributes such as force (interaction and weight), motion-constraints on objects and functional class (use and hold), whereas we propose a solution to task or manipulation action classification based on the grasp information, motion-constraints, and object class. 

\begin{figure*}[!t]
  \centering
  \includegraphics[scale=0.7]{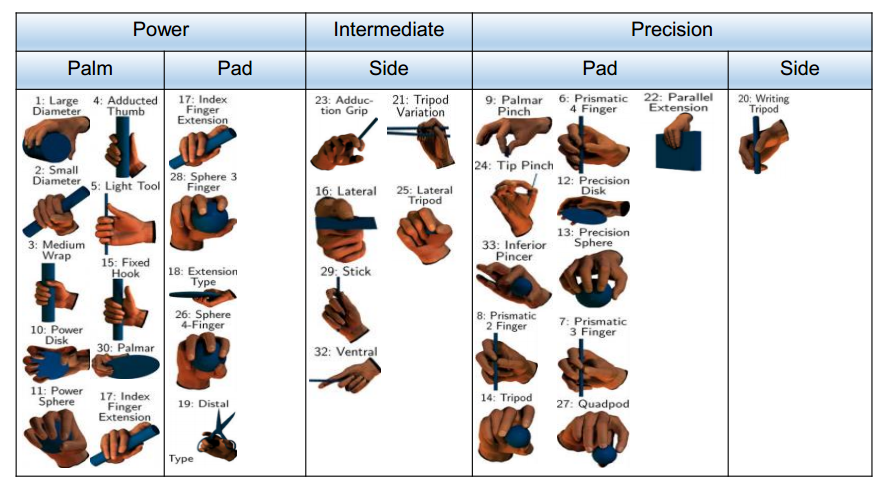}
  \caption{Coarse grasp categorization based on grasp taxonomy \cite{feix} where power, precision, and intermediate are grasp types and palm, side, and pad are opposition types.}
  \label{grasp}
\end{figure*}

Unlike the works of \cite{yezhou} and \cite{haptics}, we consider fine-grained and physically interpretable action categories, also including object information. For instance, we consider the manipulation action of towel wiping and cloth wiping as two different tasks whereas Feix et al. \cite{haptics} consider it as a single task. We believe that manipulation actions need to be classified at such a finer level to be able to serve the purpose of recognition of everyday manipulation actions and transferring complex task capabilities to the robotic manipulations. We differentiate between object manipulation actions, focusing on the functional property of an object. Thus, we demonstrate that information related to grasp, objects, and their motion-constraints are useful in achieving high recognition accuracy for a large set of action classes in an everyday manipulation action dataset. Our work \cite{kartik} considers fine-grained recognition of object manipulation actions using coarse-grained grasp attributes at instance level modeling of manipulation actions. However, in this work we also perform sequence level modeling and coarse-level action recognition of object manipulation actions. Also, we perform fine-grained action recognition based on fine-grained grasp attributes. 

The important aspects of our work include: a) A compact representation of the grasp and motion-constraints using some popular and some contemporary schemes. b) Demonstrating the usefulness of information from coarse-grained and fine-grained grasp attributes as well as motion-constraints for fine-grained and coarse-grained action recognition. c) A differential experimental analysis involving subsets of grasp and motion-constraints features, to provide more insights on the usefulness of grasp information alone, motion-constraints information alone, and grasp and motion-constraints based information together for intended classification problem. d) Comparisons between Instance and Sequence level modeling of object manipulation actions using fine-grained grasp information. e) An extensive experimental evaluation using different contemporary multi-class and binary classifiers (with a multi-class voting strategy), which also serves as a useful comparative study of popular classifiers for the manipulation action recognition problem. This analysis also helps to demonstrate that different classification frameworks, largely arrive at a consensus with respect to our hypothesis about using grasp and motion-constraints for fine-grained action classification. We demonstrate our results on a large Yale Human Grasping dataset \cite{yale} which involves various tasks on different objects. 


\section{Attributes}
\label{proposed work}
In this work, the attributes which we consider for recognition of manipulation actions, include object information, grasps, and motion-constraints of objects. 

\subsection{Object}
The object name (or corresponding symbols) serves as a simple string data on the information on the name of the object. As we want to perform classification of actions based on the grasp and motion-constraints information of the known object, we use the object name in the feature representation of an instance such as $towel$, $paper$, $bottle$, $pen$ etc. 

\subsection{Grasp attributes}
We propose to use coarse and fine level categorization of grasp types. Rest of grasp attributes have been illustrated in terms of grasped dimension, opposition type.

\subsubsection{Coarse grained grasp categorization}
There are large number of grasp taxonomies available based on earlier research on grasp types. Grasp types have also been classified at coarser and finer level (e.g. \cite{feix}), with grasp type as Power, Precision and Intermediate grasps at coarser level, as also discussed in \cite{feix}. Both fine level and coarse level grasp categorization are quite popular but the coarse level grasp categorization is relatively simple. We note that our assumption about the availability of grasp attributes, is considered to be more suited for the coarse level attributes than the finer level ones, as the latter are arguably, more difficult to estimate. Figure \ref{grasp} illustrates coarse and fine level grasp categorization for 33 grasp types specified in Feix et al. \cite{feix}.

At the coarse level, each grasp can be classified by its need for precision or power to be properly executed. The differentiation is very important, and the idea has influenced many previous studies. In the power grasp, there is a rigid contact between the object and the hand that infers all the motion for the object is based on the human arm. For the precision grasp, the hand is able to perform intrinsic movements on the object without having to move the arm. In the third category i.e. Intermediate grasp, characteristics of power and precision grasps are present in roughly the equal proportion. We demonstrate that such a coarse division among grasp attributes is also useful for the purpose of manipulation action recognition. 

\subsubsection{Fine grained grasp categorization}
As emphasized previously, grasp can also be categorized at finer level with 33 grasp types \cite{feix} as illustrated in Fig. \ref{grasp}. We use this finer level of grasp classification to compare the action recognition rates with the coarser level of grasp categorization to understand more accurately how useful the finer represent ion is to get more detailed information of grasp type in the context of manipulation action recognition.  

\subsubsection{Opposition type}
Apart from grasp type, we further use three basic directions relative to the hand coordinate frame, as illustrated in Figure \ref{grasp}, for 33 grasp types \cite{feix}. These are the directions in which, the hand can apply forces on the object to hold it securely. Pad Opposition occurs between hand surfaces along a direction generally parallel to the palm. Palm Opposition occurs between hand surfaces along a direction generally perpendicular to the palm. Side Opposition occurs between hand surfaces along a direction generally transverse to the palm.

Opposition type mainly contains the information about the direction of grasp of the object whereas Grasp type contains the information about the force on the object. Both opposition type and grasp type consists of complementary information.

\subsubsection{Grasped dimension}
\begin{figure}[!h]
  \centering
  \includegraphics[width=8.3cm, height=4.6cm]{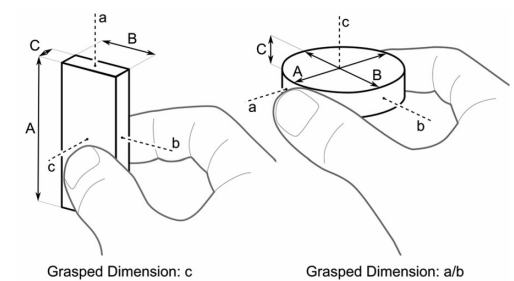}
  \caption{(Reproduced from \cite{feixobject}) Grasped dimensions for cuboid and round objects. For the round object grasp opening could be along both $a$ and $b$ dimensions.}
  \label{graspdim}
\end{figure}

In addition, we also employ grasped dimension as another feature for representation, which signifies the specific dimensions (sides) of the object along which the object is grasped. For instance, a knife needs to be grasped along the blade to be able to be used for cutting purpose. We use the grasped dimension stated in \cite{feixobject} as the part of the object that lies between the fingers when grasped. The values are from the set $(a, b, c)$ to indicate which axes best determine the hand opening. Here $a$ is along the longest object dimension and $c$ is along the shortest dimension. An example is illustrated in Figure \ref{graspdim}. The grasped dimension contains crucial information about handling of the object. It gives a spatial relationship between the human hand and object.

\subsection{Motion-Constraints on object being manipulated}
Depending on the task (and also the object properties), an object is only allowed to translate and rotate in certain directions in order to successfully complete the task. In order to categorize motion-constraints for manipulation action, each of the three axes is assigned a symbol for the motion-constraints as abbreviated in Table \ref{table:constraint}. Thus, the resultant attribute can be represented as a string with three characters (symbols). Moreover, not all the combinations for three axes (i.e. $4^3$ combinations) are practically valid, and only a set of 20 possible relative motions between two rigid bodies specified in \cite{morrow}, \cite{morris}, are used. The nomenclature defines the relationship between the object and the environment (a fixed reference frame). Table~\ref{table:constraint} illustrates the symbols used for the motion-constraints along each axes of the object being manipulated by human hand to show whether motion for the object along an axis is unconstrained or allows translation/rotation or fixed. 

\begin{table}[!h]
  \begin{center}
    \begin{tabular}{c|c|c}
      \hline
      Symbol & Translation/Rotation & Interpretation\\
      \hline
      u & unconstrained/unconstrained & \underline{u}nconstrained\\
      t & unconstrained/fixed & \underline{t}ranslation\\
      r & fixed/unconstrained & \underline{r}otation\\
      x & fixed/fixed & fi\underline{x}ed\\
      \hline
    \end{tabular}
  \end{center}
  \caption{Each of the three axes can either be free to move (u), only allow translation (t), only allow rotation (r) or do not allow any movement around that axis (x). Motion-constraints along x, y and z axes of the object is categorized using these generalizations.}
  \label{table:constraint}
\end{table}

\section{Classification}
\label{classifiers}
\subsection{Instance level modeling of manipulation actions}
As discussed above, we represent an instance of a manipulation action using grasp label (power, precision and intermediate), opposition type (palm, side and pad), grasped dimensions of the object, object name, and motion-constraints on the object. 
\begin{table}[!h]
  \begin{center}
    \begin{tabular}{c|c}
      \hline
      Variable & Abbreviation\\
      \hline 
      Grasp Type & $\nu$ \\
      Opposition Type & $\rho$ \\
      Object & $\kappa$ \\
      Grasp Dimension & $\chi$ \\
      Motion-Constraints & $\alpha$ \\
      \hline
    \end{tabular}
  \end{center}
  \caption{Abbreviations used for different grasp and motion-constraints attributes.}
  \label{abrrev}
\end{table}
To represent an instance $i$, we concatenate these string data abbreviated in Table \ref{abrrev} to form a feature vector $x_i$.
\begin{equation}
\label{eqinst}x_i=[\nu, \rho, \kappa, \chi, \alpha ]^T\end{equation}

During our experimentation for differential analysis, i.e. to see the effect of individual attributes or their subsets, we define the instance $x_i$ by removing one or more attributes from the representation in equation \ref{eqinst}.


\subsection{Sequence level modeling of manipulation actions}
In addition to action instances, we also model each manipulation action sequence of instances where grasp types are changed within each sequence. We use only those sequences for evaluation where action sequence consists of atleast two instances. For each instance we have grasp type information and the object information. We then take fine level grasp information and accumulate the number of each grasp types that fall into each sequence to represent that sequence of action. The representation is similar to the histogram of visual words representation using bag-of-words (BOW) model, where different visual words are the clusters (estimated using clustering techniques such as K-Means clustering) for all the feature vectors from the data. In BOW, the histogram feature representation for a sample is estimated by accumulating the number of features falling in each cluster. In context of our approach, the visual words are the 33 fine-grained grasp types. Each sequence of action is finally represented by 34 dimensional feature vector as in total there are 33 fine level grasp types and one object label.

\subsection{Coarse level classification of manipulation actions}
We also perform coarse level classification for each instance at the force level i.e., weight and interaction and motion-constraint level \cite{feixtask} instead of the manipulation action label (i.e. fine-grained manipulation action recognition) of the instance. The force property specifies what type of force is necessary to complete the task. Since, the forces required can be complex and difficult to discern visually, we use a simplified description that still provides useful information about the task. Specifically, we assign a value of either “weight” or “interaction”. We assign “weight” if the grasp force is closely related to lifting the object. This can be the case for tasks other than object transport, such as using a drill. In that case, the dominating force requirement is to lift the drill, squeezing the trigger usually needs less force. In the
second category, “interaction,” the grasp force is determined by factors other than object weight, usually through the interaction with the environment. There are two main mechanisms for this decoupling: the weight of the object is supported by the constraints, making the force needed to move the object less than would be required to lift the object (such as opening a drawer or door); or when the interaction force is primarily intended to apply a force through the object, such as is done when scrubbing with a sponge
(where the force needed to lift the sponge is much less than
the force needed to scrub effectively).

\subsection{Classification models}
As to our knowledge, there is no other work related to grasp and motion-constraints attributes for fine-grained classification of manipulation actions. Hence, we take this opportunity to provide classification results using various contemporary classification frameworks. These include multi-class decision forests, multi-class neural networks and multi-class classifiers constructed from binary classifiers. Such methods include locally deep support vector machines, support vector machines, binary boosted decision tree, and binary neural networks. We briefly discuss these below.

Multi-class decision forests \cite{decisionforest} and binary boosted decision trees \cite{boosted}, are extensions of decision tree based classifiers. A decision forest is an ensemble model that very rapidly builds a series of decision trees, learning from labeled data. Decision trees subdivide the feature space into regions with largely the same label. These can be regions of consistent category or of constant value, depending on whether we are doing classification or regression. Boosted decision trees avoid overfitting by limiting how many times they can subdivide and how few data points are allowed in each region. 

In both multi-class and binary neural networks which we use, input features are passed forward (never backward) through a sequence of layers before being turned into outputs. In each layer, inputs are weighted in various combinations, summed, and passed on to the next layer. This combination of simple calculations results in the ability to learn non-linear class boundaries and data trends.

Support vector machines (SVMs) \cite{libsvm} find the boundary that separates classes by as wide a margin as possible. When the two classes cannot be linearly separated, one can use kernel transformation to project the data into higher dimension, wherein classes may be arguably more separable. Two-class locally deep SVM is a non-linear variant of SVM proposed in Jose et al. 
\cite{jose}.

\begin{figure*}[t]
  \centering
  \includegraphics[scale=0.57]{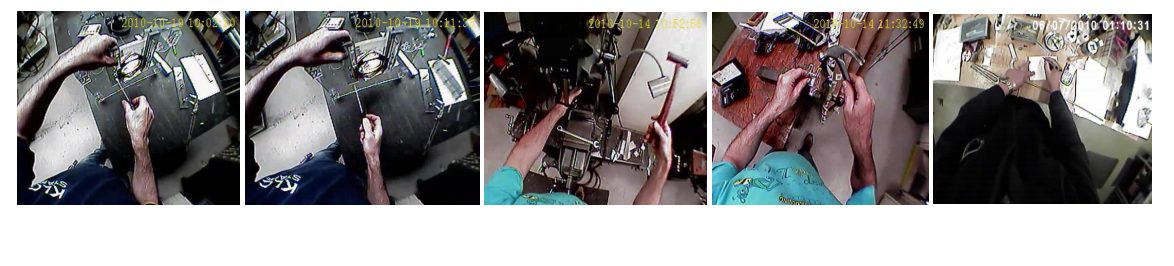}
  \caption{Sample frames of Yale human grasping dataset depicting variation in grasps for the objects - screwdriver, hammer and pen.}
  \label{dataset}
\end{figure*}

As indicated above, one can perform a multi-class classification using binary classifiers. Typically, such schemes use one-vs-one classification, and construct one classifier per pair of classes. This approach requires the modeling of $N(N-1)/2$ classifiers, where $N$ denotes the number of classes. During the testing stage, the test sample receiving the most votes from any class label is assigned that label. In the event of a tie (among two classes with equal number of votes), the label selection is based on the class with the highest aggregate classification confidence by summing over the pair-wise classification confidence levels computed by the underlying binary classifiers.

\section{Experiments and results}
\label{results}
As mentioned earlier, we evaluate our proposed hypothesis on the Yale human grasping dataset \cite{yale} consisting of everyday manipulation actions. We also emphasize that, to our knowledge, this is the only publicly available dataset which considers such a large set of everyday manipulation action in an unstructured environment. We evaluate the classification using different multi-class classifiers and binary classifiers with one-vs-one multi-class voting scheme to model the grasp and motion-constraints information. We also provide some differential analysis over the attributes, to study their effect on classification.

\subsection{Yale human grasping dataset}
This dataset consists of large annotated videos of housekeeper and machinist grasping in unstructured environments. The full dataset contains 27.7 hours of tagged video and represents a wide range of manipulative behaviors spanning much of the typical human hand usage. It involves total of 455 distinct manipulation actions (excluding holding actions and the action without proper grasp information) performed by two machinists and two housekeepers with 6188 action instances. Some example images from this dataset are illustrated in Figure \ref{dataset}, involving different grasps on some of the common objects like screwdriver, hammer, and pen. The videos are acquired by a head mounted camera on each subject. All subjects have normal physical ability, are right handed, and have been able to generate at least 8 hours of data. The labels for each of the task attributes, grasp attributes and object attributes are available with the dataset itself. This dataset is annotated by the raters experienced in the domain. We use different attributes such as grasp, motion-constraints, object name as features, and task attributes, which are available in the dataset as label for each action instance.

\subsection{Experimental settings}
We evaluate the proposed hypothesis on Yale human grasping dataset using two fold cross validation scheme where 50\% of instances of each action associated with an object are used for training purpose and rest are used for testing purpose. As it is not necessary that all the actions performed by one machinist/housekeeper are performed by other machinist/housekeeper, we do not use a cross subject evaluation here. We remove the instances of task for which raters are not able to annotate any grasp information. Also, the task $holding$ is trivial as a manipulation action so we get rid of those instances too. We ultimately concatenate the object and task string data for each instance to get manipulation action labels. These labels serves as our manipulation actions as the goal for us is to classify which action is being done using a particular object. Finally, we have 455 different manipulation actions after the cleaning of dataset for our purpose with a total of 6188 manipulation action instances.

\subsection{Results and discussion}
\subsubsection{Fine-Grained action recognition based on coarse-level grasp, motion-constraints, and rest grasp attributes}
We first provide recognition results (Table \ref{resultcombined}) using only the object and grasp attributes (without motion-constraints). These results indicate that even partial grasp information is quite useful enough to classify a large set of 455 complex manipulation actions. This information is useful to understand that even with methods to recognize grasp types at much coarser level, one can distinguish between the complex manipulation actions to some extent. Table \ref{resultcombined} also shows differential recognition rates based on the individual grasp attributes (grasp type, grasped dimension, and opposition type). From these results, we can infer that opposition type contributes relatively more to the recognition results. However, most of the classifiers agree that the combined attributes do perform better than individual ones (as expected). In general, this clearly highlights that grasp attributes indeed provide quite useful information for manipulation action recognition, and the fact that we are using a large dataset, support such a hypothesis. Even with a large set of action classes, we are able to differentiate tasks based on the object and grasp information at a rate of 0.7085. Also in above experiments one can observe that, all the classifiers perform similar, but neural networks perform somewhat better.

\begin{table*}[!t]
  \begin{center}
    \begin{tabular}{|l|p{1.5cm}|p{1.8cm}|p{1.7cm}|p{1.7cm}|}
      \hline
      Classifier & Grasp Type (PIP) & Opposition Type & Grasped Dimension & Grasp Information(All)\\
      \hline
      \hline
      Multi-class decision forest & 0.6460 & 0.6532 & 0.6508 & \bf{0.6966}\\
      Multi-class neural network & 0.6810 & 0.6820 & 0.6474 & \bf{0.6929}\\
      Locally deep SVM (Binary) & 0.6688 & 0.6908 & 0.6677 & \bf{0.6943}\\
      SVMs (Binary) & 0.6789 & \bf{0.6912} & 0.6644 & 0.6854\\
      Neural network (Binary) & 0.6973 & 0.7041 & 0.6508 & \bf{0.7085}\\
      \hline
    \end{tabular}
  \end{center}
  \caption{Action recognition results (top accuracies across columns in bold) for two fold cross validation evaluation based on different grasp attributes using different multi-class and binary classifiers.}
  \label{resultcombined}
\end{table*}

\begin{table}[!h]
  \begin{center}
    \begin{tabular}{|l|c|}
      \hline
      Classifier & Avg. accuracy\\
      \hline
      \hline
      Multi-class decision forest & 0.8235 \\
      Multi-class neural network & 0.8262 \\
      Locally deep SVM (Binary) & 0.8445\\
      Support vector machine (Binary) & 0.8408 \\
      Neural network (Binary) & 0.8327 \\
      \hline
    \end{tabular}
  \end{center}
  \caption{Action recognition results for two fold cross validation evaluation based on motion-constraints attributes using different multi-class and binary classifiers.}
  \label{resultcons}
\end{table}

We next provide, in Table \ref{resultcons} the recognition results with objects and motion-constraints alone (without grasp attributes), and in Table \ref{result2}, results with all attributes. These results indicate that motion-constraints appears to help the manipulation action classification, much more than grasp information. However, in Table \ref{result2}, one can notice that most classifiers agree that grasp attributes further improves the overall classification up to some extent. Below, we take a closer look at the difference between grasp and motion-constraints, considering certain specific classes. 

\begin{table}[!h]
  \begin{center}
    \begin{tabular}{|l|c|}
      \hline
      Classifier & Avg. accuracy\\
      \hline
      \hline
      Multi-class decision forest & 0.8310 \\
      Multi-class neural network & 0.8388 \\
      Locally deep SVM (Binary) & 0.8293\\
      Support vector machine (Binary) & 0.8150 \\
      Neural networks (Binary) & 0.8446 \\
      \hline
    \end{tabular}
  \end{center}
  \caption{Action recognition results for two fold cross validation evaluation based on the grasp and motion-constraints attributes using different multi-class and binary classifiers.}
  \label{result2}
\end{table}

The failure cases to the action recognition based on grasp are mainly of the objects which do not have any rigid structure. Such objects do not have a particular way of handling to complete an action, for e.g. towel, paper etc. The reason for lower recognition rates for manipulation actions using these objects based on grasp information is the non-rigid structure of the objects. Out of 6188 total action instances 19\% of total instances i.e. 1189 instances consists of manipulation actions using towel. These object manipulation actions still are able to achieve better recognition rates based on the motion-constraints attributes as most of the actions based on these objects allow limited degree of freedom for the motion of object, for e.g. $cloth/towel~wiping$ on plane surface does not usually consists of rotation along two axes and translation along one axis. 

We perform another experiment to support this hypothesis, by removing instances of object - towel, cloth, and paper (where $towel$ constitute 19\% of instances of whole dataset). In Table \ref{impconc}, we provide the results for this experiment. One can clearly notice in the earlier recognition results (across Tables \ref{resultcombined} and \ref{resultcons}), the difference between the results with grasp and motion-constraints is of the order of 10\% to 15\%. However, after removing the ``non-informative'' classes from the grasp perspective, one can observe that the classification using grasp attributes has also improved dramatically. While, the motion-constraints still contribute more for the recognition, the difference between recognition using grasp and motion-constraints is now reduced to 2\% - 3\%. Moreover, combining grasps and motion-constraints consistently improves performance over their individual ones. Such a differential analysis highlights that while motion-constraints are generally useful for recognition, grasp attributes are also important, except for a small fraction of classes.

\begin{table}[h]
  \begin{center}
    \begin{tabular}{|l|l|l|l|}
      \hline
      Classifiers & Grasp & Motion-Constraints & Both\\
      \hline
      \hline
      Multi-class decision forest & 0.7840 & 0.8022 & \bf{0.8088}\\
      Multi-class neural networks & 0.7913 & 0.8166 & \bf{0.8378}\\
      Locally Deep SVM (Binary) & 0.7876 & 0.8236 & \bf{0.8286} \\
      SVM (Binary) & 0.7819 & 0.8205 & \bf{0.8495}\\
      Neural Networks (Binary) & 0.8045 & 0.8218 & \bf{0.8318}\\
      \hline
    \end{tabular}
  \end{center}
  \caption{Action recognition results (top accuracies across columns in bold) for two fold cross validation evaluation using different classifiers after removing instances involving objects - towel, cloth, and paper.}
  \label{impconc}
\end{table}

Such an inference is vital considering that the motion-constraints information i.e. degrees of freedom of object for the manipulation action, is relatively difficult to understand from the manipulation actions as compared to grasp information at a coarser level, using existing methods. Thus, one can appreciate that such coarse grasp information (which is easier to compute) can still prove useful to the manipulation action recognition. 

The above analysis also serves to provide a comparison among different contemporary classifiers, for the current task involving categorical features provided in Yale human grasping dataset. We note that in majority of the cases binary neural network yields high classification accuracies. In addition, SVMs and multi-class neural networks also perform well, and often provide close to highest accuracies. It is also observed that the decision forest classifiers yield relatively low classification rates. 

\subsubsection{Fine-Grained action recognition based on fine-level grasp, motion-constraints, and rest grasp attributes}
Recently, there has been considerable amount of research for the fine-grained recognition of grasps such as \cite{rogez}, \cite{cai2015}. Intuitively, this problem is more challenging than the coarse grained grasp recognition due to obvious reason of classifying at much finer level. As we focus on action recognition, we consider here the reverse problem which uses fine-grained grasp information (Table \ref{resultcombined2}). For this, 33 fine-grained grasp types as illustrated in Fig. \ref{grasp} are used. The other grasp attributes mentioned in Table \ref{resultcombined2} constitute of opposition type and grasped dimension.  
\begin{table*}[!h]
  \begin{center}
    \begin{tabular}{|l|p{1.3cm}|p{1.7cm}|p{1.7cm}|p{2.2cm}|}
      \hline
      Classifier & Grasp type(33) & Grasp type(33) and other grasp attributes & Grasp type(33) and motion-constraints & Grasp (fine), Grasp (coarse), other grasp attributes \& motion-constraints\\
      \hline
      \hline
      Multi-class decision forest & 0.7102 & 0.7197 & \bf{0.8378} & 0.8327\\
      Multi-class neural network & 0.7703 & 0.7740 & 0.8805 & \bf{0.8809}\\
      Locally deep SVM (Binary) & 0.7224 & 0.7288 & 0.8517 & \bf{0.8548}\\
      SVM (Binary) & 0.7367 & 0.7187 & \bf{0.8296} & 0.8185\\
      Neural network (Binary) & 0.7404 & 0.7397 & \bf{0.8541} & 0.8538\\
      Ensemble & 0.7455 & 0.7431 & \bf{0.8585} & 0.8510\\

      \hline
    \end{tabular}
  \end{center}
  \caption{Action recognition results (top accuracies across columns in bold) for two fold cross validation evaluation based on fine level 33 grasp types and combining them with rest grasp attributes, motion-constraints using different multi-class and binary classifiers.}
  \label{resultcombined2}
\end{table*}

There are quite a few interesting observations to note from these experiments. One is with both fine grasp information and fine grasp with rest of grasp attributes information, the recognition accuracy is nearly equal. Columns 2 and 3 in Table \ref{resultcombined2} indicates that the finer level of grasp classification substitutes for the information imbibed in the other grasp attributes such as grasp dimension, opposition type. This fact is further justified when recognition rates are nearly equal for cases with and without rest of grasp attributes and along with motion-constraints information (columns 4 and 5 in Table \ref{resultcombined2}). 

Apart from that, by comparing Table \ref{resultcombined} and \ref{resultcombined2}, we note that there is a clear increment in the recognition accuracy when we just use fine-grained grasp class labels and object class labels instead of coarse-grained grasp labels and object labels for the task of fine-grained object manipulation action recognition. This result is very much expected as we are adding a finer level of information to our grasp labels. Thus, we note that we should preferably use fine-grained grasp information, if available, rather than coarse level grasp information. 

After adding motion-constraints data in our instance representation, the difference in recognition rates with coarse and fine-grained grasp information is somewhat less. This is due to the fact that motion-constraint shows a better ability to model the instances for the task of object manipulation action recognition.  
  
\subsubsection{Coarse level action recognition based on grasp and object information}
We now perform coarse level action classification using all the grasp information such as fine and coarse level grasp types, object labels and other grasp attributes (Table \ref{coarse}). Coarse level action recognition experiments are performed at force level (weight and interaction class), and motion-constraints level (20 classes).

\begin{table}[!h]
  \begin{center}
    \begin{tabular}{|l|p{1cm}|p{1.2cm}|p{1cm}|}
      \hline
      Classifiers & Fine level (Manipulation Actions) & Coarse level (Motion-Constraints) & Coarse level (Force)\\
      \hline
      \hline
      Multi-class decision forest & 0.7197 & 0.8382 & \bf{0.8394}\\
      Multi-class neural networks & 0.7740 & 0.8761 & \bf{0.8953}\\
      Locally Deep SVM (Binary) & 0.7288 & 0.8380 & \bf{0.8542} \\
      SVM (Binary) & 0.7187 & 0.7968 & \bf{0.8167}\\
      Neural Networks (Binary) & 0.7397 & 0.8340 & \bf{0.8552}\\
      \hline
    \end{tabular}
  \end{center}
  \caption{Comparison of action recognition results at different level of classification such as level of manipulation actions, motion-constraints and force (top accuracies across columns in bold).}
  \label{coarse}
\end{table}
As expected, using full grasp information, recognition is more accurate for coarse level classification than fine level classification. As, we achieve 88\% recognition accuracy at motion-constraints level (20 classes), one of the interesting observation is the interdependency between the grasp and motion-constraints information. This observation is especially important to observe that grasp based action recognition can be a good substitute to the motion-constraints based action recognition, where understanding motion-constraints for each action instance is difficult. 

A high recognition rate for the force level based on the grasp and object information again highlights the efficacy of the grasp information. It allows one to infer what level of force (weight or interaction) is applied with a specific grasp for an object. It indicates a high level understanding for actions eg. drilling requires lifting of the machine therefore requires weighted force, whereas writing with a pen requires interaction force. To transfer the manipulation capabilities to robots, such an observation is really important. 

\subsubsection{Sequence level action recognition based on fine level grasp information}
\begin{table}[!h]
  \begin{center}
    \begin{tabular}{|l|p{2cm}|p{2cm}|}
      \hline
      Classifiers & Sequence level & Instance level \\
      \hline
      \hline
      Multi-class decision forest & 0.7029 & \bf{0.7503}\\
      Multi-class neural networks & 0.7256 & \bf{0.7799}\\
      Locally Deep SVM (Binary) & \bf{0.7664} & 0.7534\\
      SVM (Binary) & \bf{0.7642} & 0.7534\\
      Neural Networks (Binary) & 0.7551 & \bf{0.7653}\\
      Ensemble & \bf{0.7846} & 0.7695\\
      \hline
    \end{tabular}
  \end{center}
  \caption{Comparison of action recognition results for sequence level and instance level classification taking actions having more than one sequence in the Yale human grasping dataset i.e. 105 actions (top accuracies across columns in bold).}
  \label{seq1}
\end{table}

Finally, we show the recognition results for sequence level modeling of action based on the fine level grasp information. Sequence level modeling is based on 34 dimensional feature vector where each feature dimension represents the count of the specific fine-grained grasp type involved in that action sequence. This type of action modeling is expected to be a better way to model complex actions where multiple types of grasp types are involved, whereas a instance level modeling would either get confused for such an action sequence as the same action sequence will be co-related to different fine grained grasp types. We only use actions having more than one sequence (thus have 105 actions out of total 455 manipulation actions) in Table \ref{seq1}, and more than five sequence (thus having 39 actions out of total 455 manipulation actions) in Table \ref{seq2}. 

In Table \ref{seq1}, we note that generally, instance level recognition results are better than the sequence level recognition results.This could be due to less training examples in sequence level case (as we just have minimum one example for training and one example for testing). To model the sequence level information, we need to have more training examples even with the approach similar to bag-of-words.

\begin{table}[!h]
  \begin{center}
    \begin{tabular}{|l|p{2cm}|p{2cm}|}
      \hline
      Classifiers & Sequence level & Instance level \\
      \hline
      \hline
      Multi-class decision forest & 0.7694 & \bf{0.7972}\\
      Multi-class neural networks & \bf{0.8055} & 0.8015\\
      Locally Deep SVM (Binary) & \bf{0.8028} & 0.7982\\
      SVM (Binary) & 0.7694 & \bf{0.7953}\\
      Neural Networks (Binary) & \bf{0.8055} & 0.7977\\
      Ensemble & \bf{0.8000} & 0.7991\\
      \hline
    \end{tabular}
  \end{center}
  \caption{Comparison of action recognition results for sequence level and instance level classification taking actions having more than 5 sequences in the Yale human grasping dataset i.e. 39 actions (top accuracies across columns in bold).}
  \label{seq2}
\end{table}

Finally, we consider those actions which have more than 5 sequences to address the issue of lesser training examples in Table \ref{seq2}. Here, the sequence level modeling performs marginally better than instance level modeling.  

\section{Conclusions}
In this paper, we present a novel approach for the recognition of everyday manipulation actions based on the grasp and motion-constraints information. We evaluate our hypothesis on large Yale human grasping dataset consisting of 455 action classes. Our results and a varied experimental analysis clearly shows that grasp information contains important clue to the everyday manipulation actions. We consider the differentiation between the functionality of the object and show that this approach for recognition has a clear advantage over the traditional methods of action recognition based on the human dynamics. Another overall advantage to this approach is that this type of action analysis is shown to work over a large set of action classes with very subtle variations in their motion dynamics. Our work indicates that considering grasp information, and object motion-constraints, one can transfer advance task capabilities to the robotics applications and modeling the human behavior in complex environment.

%
%

\bibliography{TransPaper}
\bibliographystyle{IEEEtran}









\end{document}